\documentclass[conference]{IEEEtran}
\IEEEoverridecommandlockouts
\usepackage{cite}
\usepackage{amsmath,amssymb,amsfonts}
\usepackage{algorithmic}
\usepackage{graphicx}
\usepackage{float}
\usepackage{textcomp}
\usepackage{xcolor}
\def\BibTeX{{\rm B\kern-.05em{\sc i\kern-.025em b}\kern-.08em
    T\kern-.1667em\lower.7ex\hbox{E}\kern-.125emX}}
\begin{document}

\title{Comparing Multiclass Classification Algorithms for Financial Distress Prediction}

\author{\IEEEauthorblockN{ Noopur Zambare}
\IEEEauthorblockA{\textit{Department of Mechanical Engineering} \\
\textit{Indian Institute of Technology}\\
Jodhpur, India \\
zambare.1@iitj.ac.in}
\and
\IEEEauthorblockN{ Ravindranath Sawane}
\IEEEauthorblockA{\textit{Department of Computer Science} \\
\textit{Western University}\\
Ontario, Canada \\
rsawane@uwo.ca}

}

\maketitle

\begin{abstract}
In this study, we explore how to improve the functionality of multiclass classification algorithms. We used a benchmark dataset from Kaggle to create a framework. They have been used in a number of fields, including image recognition, natural language processing, and bioinformatics. This study is focused on the prediction of financial distress in companies in addition to the wider application in multiclass classification. Identifying businesses that are likely to experience financial distress is a crucial task in the fields of finance and risk management. Whenever a business experiences serious challenges keeping its operations going and meeting its financial responsibilities, it is said to be in financial distress. It commonly happens when a company has a sharp and sustained recession in profitability, cash flow issues, or an unsustainable level of debt.

\end{abstract}


\section{Introduction}
    \subsection{Background}
    Financial distress refers to a state in which a company faces considerable challenges in meeting its financial obligations. Early indications of financial problems might help proactive actions like restructuring, obtaining more finance, or putting cost-cutting measures into place. 
        
    We used a wide range of supervised learning algorithms, such as Decision Trees, Random Forest Classifiers, and Naive Bayes, to create the framework. We intend to study the potential advantages and performance enhancements that can be achieved by combining supervised learning.

    \subsection{Problem Statement}
     The goal of this paper is to investigate the use multiclass classification in financial problems. The subject of interest is the application of multiclass classification for multiclass classification to predict financial distress in businesses. \newline
    By effectively resolving this problem, we want to open up the possibility of applying reinforcement learning principles to a variety of classification problems.

\section{Methodology}

    \subsection{Dataset}
    The study involves the use of a dataset gathered by Kaggle that includes different financial parameters and company characteristics. The dataset, which is accessible in CSV format, includes statistics on the company's performance as well as relevant contextual information. Using methods like label encoding, a preprocessing step is implemented to handle missing data, normalise features, and transform categorical variables. Then, training and testing sets are created from the preprocessed dataset.

    \subsection{Baseline Multiclass Classification Algorithms}
        \subsubsection{Decision Tree}
        In this algorithm, the space of features is recursively divided according to a set of criteria in order to generate a decision tree. Information gain or Gini impurity is the most widely used criterion. 
        They can handle categorical and numerical features, as well as non-linear relationships, and they can capture both. Decision trees, show a tendency to overfit the training set if they are not appropriately regularised or pruned. Overfitting can be reduced using strategies like pruning, establishing a minimum number of samples needed to split a node, or using ensemble methods.\vspace{7pt}
        
        \subsubsection{Random Forest Classifier}
        An ensemble technique called the Random Forest Classifier combines several decision trees to produce predictions. A random subset of features is taken into account at each split of each tree, which is trained on a bootstrap sample of the training data. By combining the predictions of various trees, either through majority voting or averaging, the final prediction is obtained. \vspace{7pt}

        \subsubsection{Naive Bayes}
        The Naive Bayes algorithm is a probabilistic classifier that relies on the Bayes theorem and makes the assumption that features are independent of the class. Given the input features, it calculates the probabilities of each class and chooses the class with the highest probability as the prediction.

    \subsection{Multiclass Classification Algorithms
    }
    
        \subsubsection{Defining Agent}
        The DQN class is used to represent the agent. Based on the input features given, it acts as the decision-making entity that learns to categorise the different levels of financial distress. The agent employs a method akin to the DQN, using a group of Decision Tree Classifier, Random Forest Classifier and Naive Bayes models as the Q-network. \vspace{7pt}

        \subsubsection{Defining Environment}
        In this case, the environment is the classification problem itself, which involves determining the levels of economic distress based on the given input features. The agent receives rewards from the environment as feedback, which helps it improve its classification performance. \vspace{7pt}

        \subsubsection{State Representation}
        The input features that were utilised to train the agent define the state representation. In this instance, the features Company, Time, x1, x2, x3, and x4 serve as representations of the state. These features are taken out of the data frame and sent to the classification agent as input. \vspace{7pt}
        
        \subsubsection{Setting Reward Function}
        The act() method of the DQN class contains a definition of the reward. If any of the true class labels in the y variable match the predicted action (class label), the agent is rewarded with a value of 1. If not, it is rewarded with -1. The goal of the reward system is to encourage the agent to forecast classes correctly. \vspace{7pt}
    
        \subsubsection{Selection of Action}
        The action selection method makes sure that the model chooses the best class label depending on the situation at hand and previously learnt information.
        The class labels that are available in this situation make up the action space. To determine the class for a particular input, the agent will select an action (class label) from this collection. The number of classes in the classification problem and the size of the action space are related.\vspace{7pt}
        
        \subsubsection{Training}
        Iterating through episodes and the stages in each episode are both parts of the training process. In agreement with an epsilon-greedy exploration-exploitation strategy, the agent chooses a course of action (class label). In accordance with the accuracy of its forecast, it is rewarded, and the ensemble of decision tree models is updated. For the specified number of episodes, training is ongoing. \vspace{7pt}
    
        \subsubsection{Evaluation}
        By comparing the predicted labels with the actual labels using the test data, it is feasible to assess how accurate the agent's predictions were. The calculated accuracy of the base model and the accuracy of the DQN-based agent after training are compared.
        The various metrics involved in analysis are accuracy, recall score and precision score. However, the performance of models was also analyzed using a confusion matrix. \vspace{7pt}

\begin{table}[H]
    \centering
    \caption{Results}
    \begin{tabular}{ |p{2.8cm}|p{1.5cm}|p{1.5cm}|p{1.5cm}|  }
     \hline
     \multicolumn{4}{|c|}{Comparative Analysis} \\
     \hline
     Model & Accuracy & Recall & Precision \\
     \hline
         Decision Tree & 0.98 & 0.50 & 0.50\\
         new & 0.36 & 0.28 & 0.34\\
         \hline
         Random Forest & 0.99 & 0.50 & 0.50\\
         new & 0.35 & 0.29 & 0.34\\
         \hline
         Naive Bayes & 0.99 & 0.75 & 0.67\\
         new & 0.36 & 0.28 & 0.34\\
     \hline
    \end{tabular}
        \label{tab:my_label}
\end{table}

\section{Results and Analysis}

\subsection{Comparison with Baseline Algorithms}
On the chosen benchmark datasets, the performance of the proposed framework, which incorporates Deep Q-Network with multiclass classification algorithms, is compared with that of the baseline algorithms.

\subsection{Analysis of Computational Efficiency}
\begin{itemize}
    \item The Decision Tree Classifier baseline model trains faster than the DQN-based model. Because they immediately learn decision boundaries and feature splits, decision trees can be trained quickly without iterative optimisation. The DQN-based model trains an ensemble of Decision Tree Classifiers periodically, which is more computationally expensive.

    \item The baseline model (Random Forest Classifier) takes less training time than the DQN-based model. Random forests are successful because they may construct many decision trees at once utilising parallel computing. Individual decision trees are trained using a random collection of characteristics and data samples. The DQN-based approach requires training for an ensemble of Random Forest Classifiers, which can be computationally intensive.

    \item Basic Gaussian Naive Bayes is computationally effective for training and prediction due to its simplicity. The ensemble DQN model uses Naive Bayes classifiers, however, it is more sophisticated and requires more processing.

    \item A single Naive Bayes classifier, along with its associated parameters and probability distributions, must be stored in memory by the base model (Gaussian Naive Bayes). In order to store an ensemble of Naive Bayes classifiers, which consists of various models with their unique parameters and probability distributions, the DQN model needs memory.

    \item In conclusion, compared to the DQN-based model, the baseline models are anticipated to be computationally more efficient in terms of training time, inference time, and memory use. 

\end{itemize}

\begin{figure}[htbp]
\centerline{\includegraphics[width=3in]{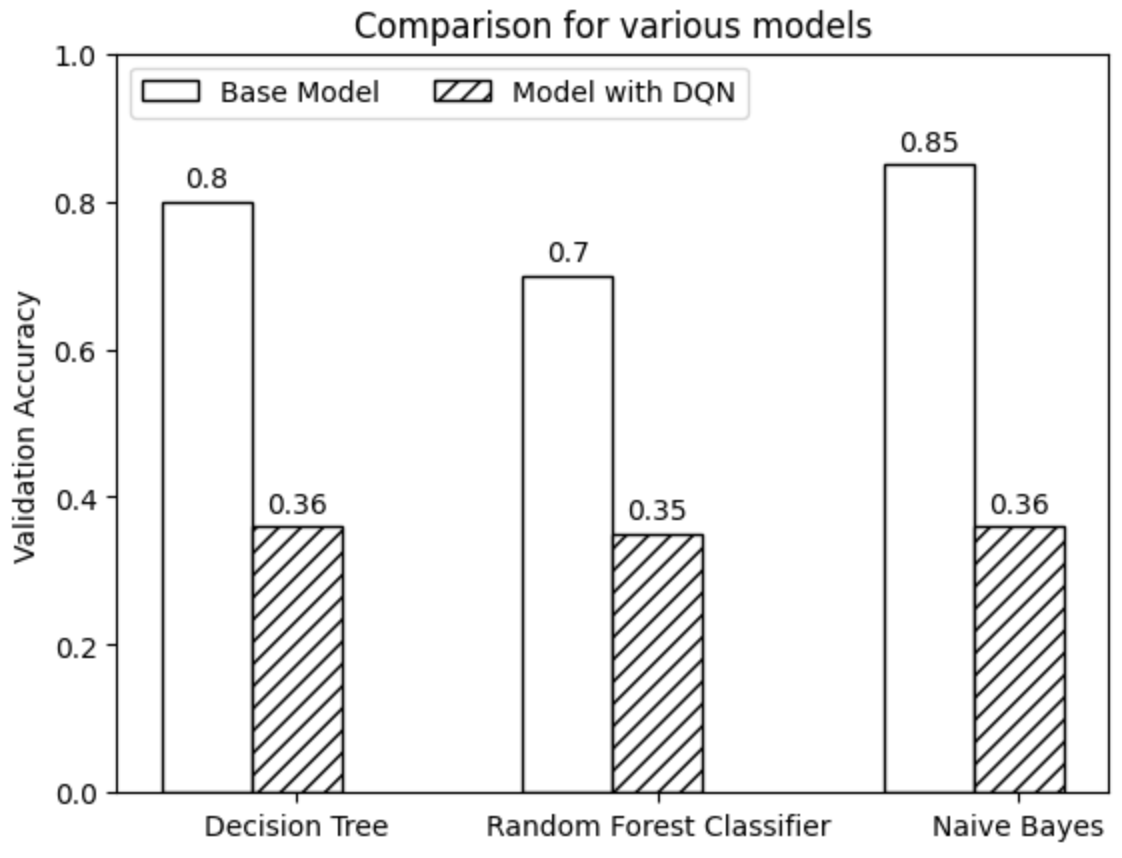}}
\caption{Comparative analysis for accuracy}

\label{fig}
\end{figure}

\begin{figure}[htbp]
\centerline{\includegraphics[width=3in]{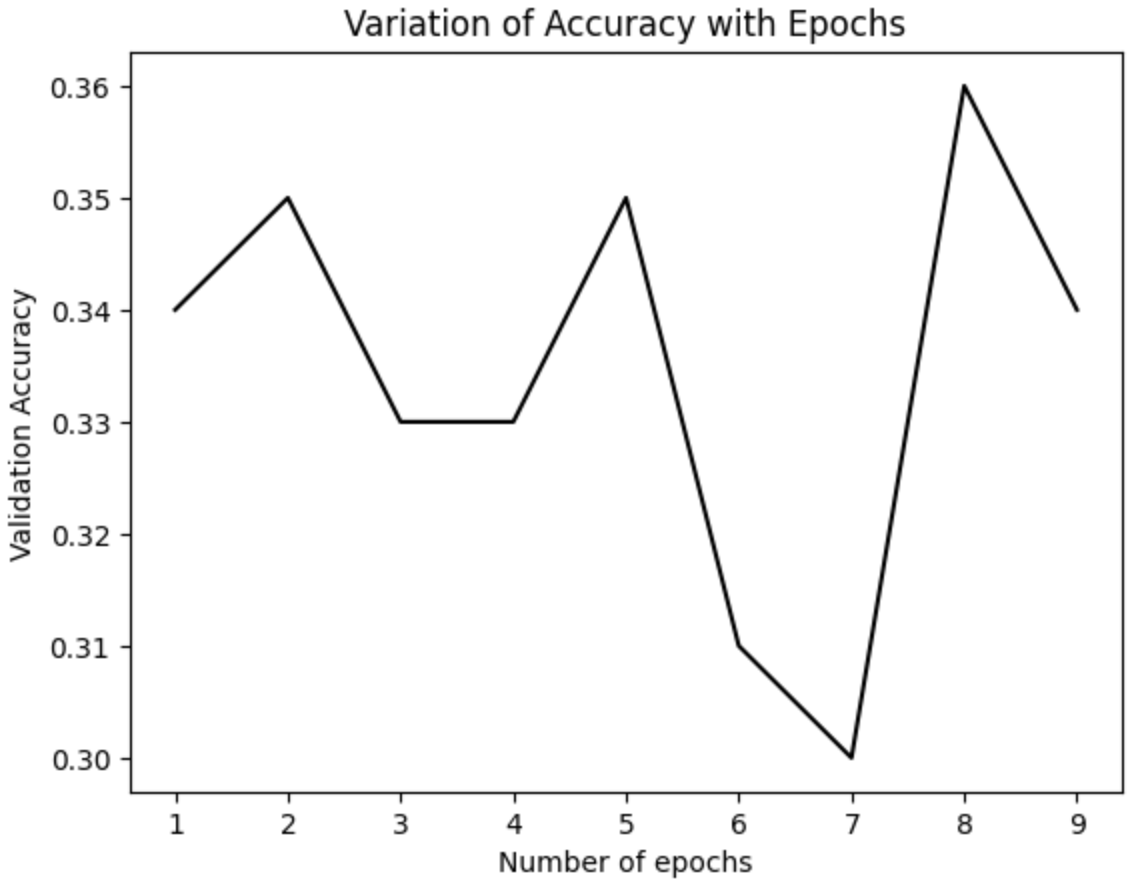}}
\caption{Variation of accuracy with epochs for Decision Tree}
\label{fig}
\end{figure}

\begin{figure}[htbp]
\centerline{\includegraphics[width=3in]{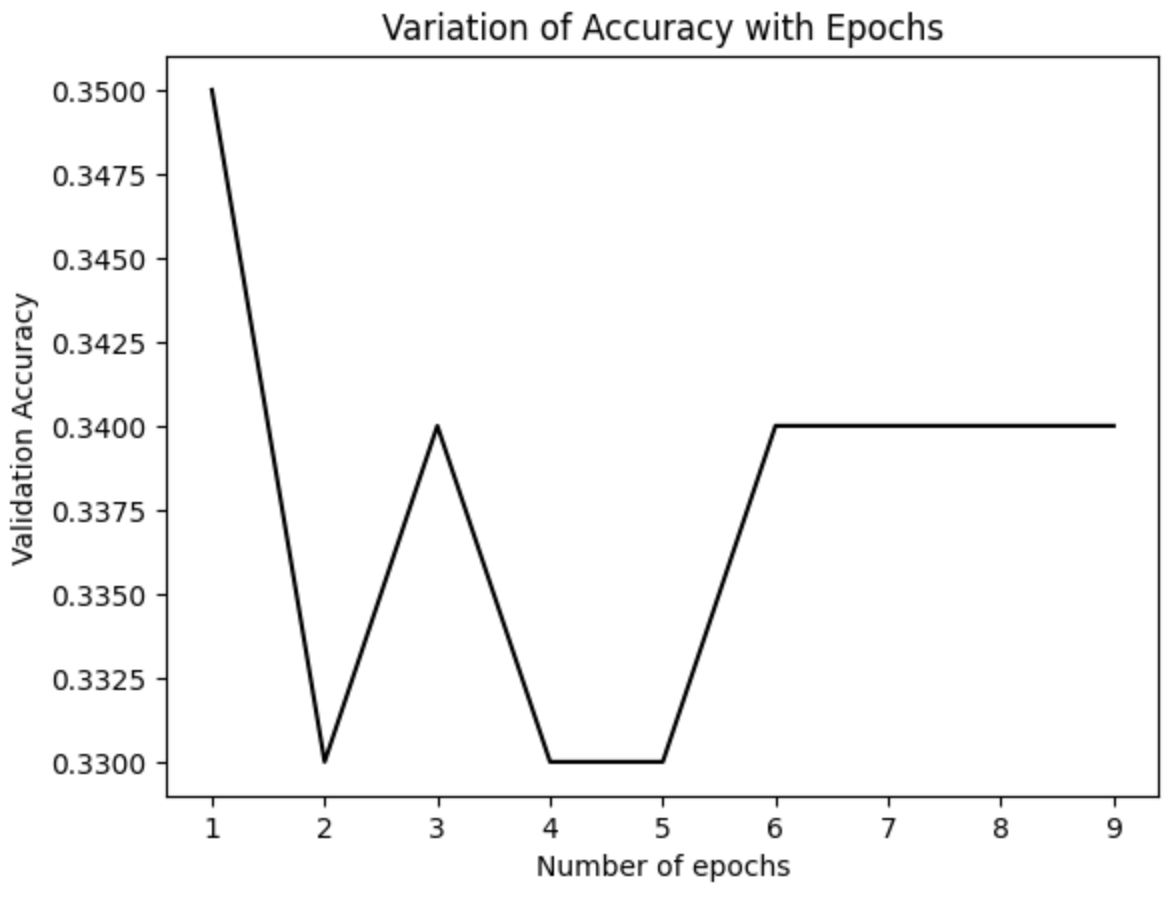}}
\caption{Variation of accuracy with epochs for Random Forest Classifier }
\label{fig}
\end{figure}

\begin{figure}[htbp]
\centerline{\includegraphics[width=3in]{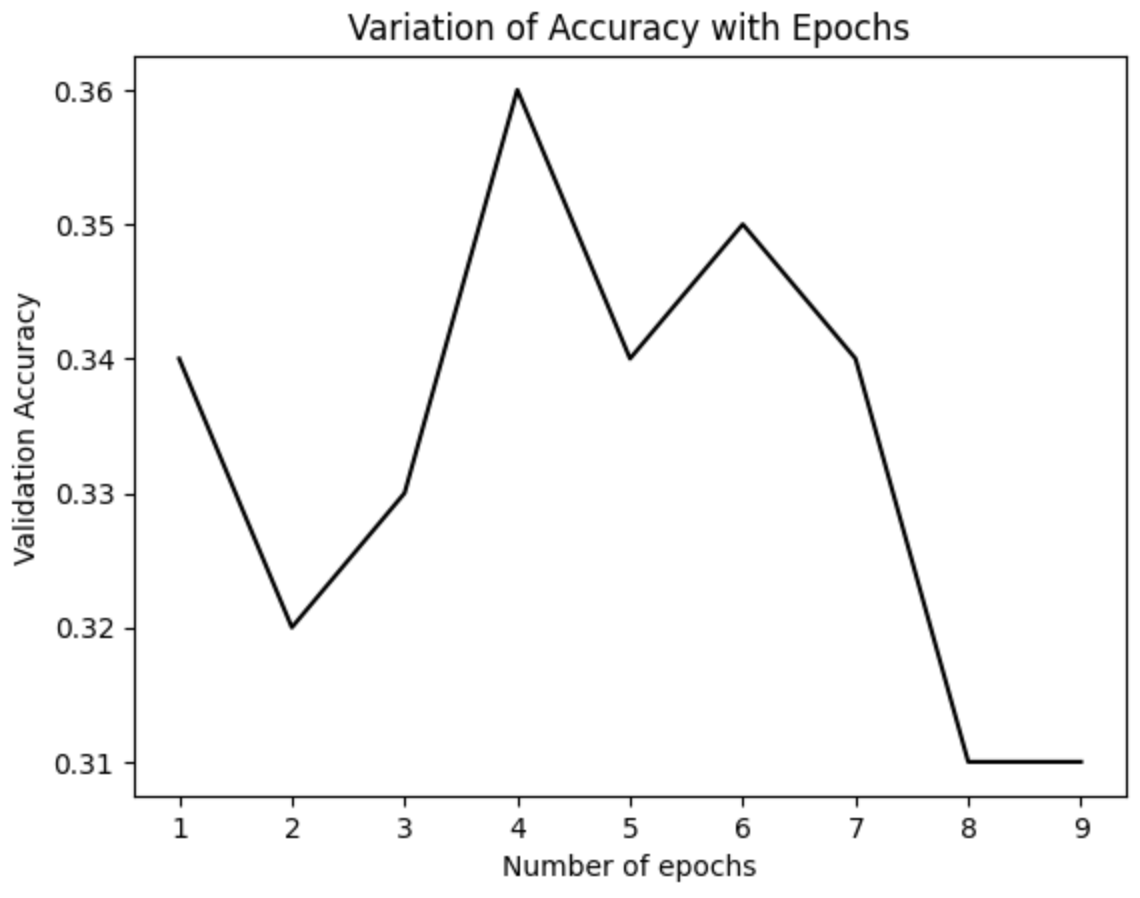}}
\caption{Variation of accuracy with epochs for Naive Bayes}
\label{fig}
\end{figure}

\section{Discussion}

    \subsection{Advantages}
    Multiclass classification methods that incorporate Deep Q-Network (DQN) have various benefits and provide special capabilities to the task. Benefits involve : 
        \begin{itemize}
            \item Handling Complex Decision-Making \newline
            The Q-function captures complicated choice settings in complex environments by giving greater values to state-action combinations that result in better outcomes. DQNs can now manoeuvre through complex scenarios with a wide range of potential actions and states thanks to this. Through successive improvements that reduce the discrepancy between the predicted Q-values and the true rewards received through contact with the environment, the DQN learns the Q-function.

            \item Adaptability to Dynamic Environments \newline
            The DQN adjusts its Q-function as the agent learns and interacts with the environment. This lets the DQN adapt its decision-making technique to external changes, guaranteeing it can still make successful decisions.

            \item Handling Imbalanced Datasets \newline
            The uneven distribution of various classes or states is referred to as an imbalanced dataset. This might happen in the setting of DQNs when some states or behaviours are observed more frequently than others. Due to the DQN's potential limited exposure to specific state-action combinations, imbalanced datasets can cause biases in the estimate of the Q-function. Techniques like experience replay and prioritised replay can be used to remedy this. By ensuring that the DQN samples experience various states and actions consistently, these techniques lessen the negative effects of imbalanced datasets on learning.
            
            \item Real-time Classification \newline
            Real-time categorization requires rapid, effective action in situations when timeliness is critical. By adopting methods that give consideration to trade-offs between exploitation and exploration, DQNs can be trained to take decisions in the present. Epsilon-greedy exploration is one such technique that aids the agent in striking a balance between selecting activities that are known to be beneficial (exploitation) and investigating novel behaviours to find potential enhancements (exploration). DQNs may take decisions rapidly while being able to learn from fresh experiences by striking this equilibrium.

        \end{itemize}
    
    \subsection{Limitations}
    \begin{itemize}
        \item Large memory requirements \newline
        Especially when employing experience replay, which includes storing and sampling from a significant replay buffer, DQN often needs a lot of RAM. 
        \item Curse of dimensionality \newline
        Finding the most effective measures and achieving efficient convergence can be more difficult when the DQN training and learning process is impacted by the curse of dimensionality. Consequently, DQN's ability to do multiclass classification well may be constrained by its ability to handle significant feature spaces.
        \item Limited generalization to new classes \newline
        It often acquires policies unique to the classes found in the training set. They are efficient at handling well-known classes, but they have a limited ability to generalise to unfamiliar or new classes. In dynamic classification contexts where new classes continually emerge, the technique is less adaptive since incorporating new classes into the model often requires retraining or considerable fine-tuning.
    
    \end{itemize}

    \subsection{Future Scope}
    Future prospects are promising when Deep Q-Network is incorporated into multiclass classification algorithms. Such as Transfer Learning and Knowledge Transfer, Real-time Classification, Hierarchical Multiclass Classification, Adaptive Learning and Dynamic Feature Selection, and many others.

\section{Conclusion}
The study uses multiclass classification to show the significance of using DQN for financial distress prediction in businesses. The study's findings may help businesses, investors, and financial institutions make informed decisions and take preventive action to reduce the risks associated with the financial crisis. 

Possible reasons for less accuracy by the DQN model than the base model :
\begin{itemize}
    \item The classifier for the base model is trained directly on the labelled training data using a traditional supervised learning methodology. In a single step, it learns the probability distributions and class boundaries from the data. While the DQN model iteratively changes its ensemble of classifiers based on the rewards it receives from the environment, it is trained using a reinforcement learning methodology. This recurrent training procedure may generate noise and instability, resulting in less accurate convergence.
    
    \item While lowering bias and variance can help ensembles perform better, they also add to the complexity and risk of inconsistencies across the various models. Lower accuracy may be the consequence if the ensemble is unable to fully capture the underlying patterns and relationships.
\end{itemize}

\end{document}